\documentclass[sigconf]{acmart}

\copyrightyear{2024}
\acmYear{2024}
\setcopyright{acmlicensed}
\acmConference[CIKM '24] {Proceedings of the 33rd ACM International Conference on Information and Knowledge Management}{October 21--25, 2024}{Boise, ID, USA.}
\acmBooktitle{Proceedings of the 33rd ACM International Conference on Information and Knowledge Management (CIKM '24), October 21--25, 2024, Boise, ID, USA}
\acmISBN{979-8-4007-0436-9/24/10}
\acmDOI{10.1145/3627673.3679559}
\usepackage{appendix}
\usepackage{hyperref}
\usepackage{caption}
\usepackage{tikz}
\usepackage{subfigure}
\usepackage{bbm}

\usetikzlibrary{decorations}
\usetikzlibrary{decorations.pathmorphing}
\usetikzlibrary{decorations.pathreplacing}
\usetikzlibrary{decorations.shapes}
\usetikzlibrary{decorations.text}
\usetikzlibrary{decorations.markings}
\usetikzlibrary{decorations.fractals}
\usetikzlibrary{decorations.footprints}
\usetikzlibrary{arrows.meta}

\definecolor{burgundy}{rgb}{0.5, 0.0, 0.13}
\definecolor{lavande}{RGB}{176,224,230}
\definecolor{lemonchiffon}{RGB}{255,250,205}

\AtBeginDocument{%
  \providecommand\BibTeX{{%
    \normalfont B\kern-0.5em{\scshape i\kern-0.25em b}\kern-0.8em\TeX}}}

\begin{document}

\title{Retrieval Augmented Deep Anomaly Detection for Tabular Data}


\author{Hugo Thimonier}
\email{hugo.thimonier@lisn.fr}
\orcid{0000-0003-4762-477X}
\affiliation{%
  \institution{Universit\'e Paris-Saclay, CNRS, CentraleSup\'elec, Laboratoire Interdisciplinaire des Sciences du Numérique}
  \city{Gif-sur-Yvette}
  \country{France}
  \postcode{91190}
  }

\author{Fabrice Popineau}
\orcid{0000-0002-2941-9046}
\email{fabrice.popineau@lisn.fr}
\affiliation{%
  \institution{Universit\'e Paris-Saclay, CNRS, CentraleSup\'elec, Laboratoire Interdisciplinaire des Sciences du Numérique}
  \city{Gif-sur-Yvette}
  \country{France}
  \postcode{91190}
  }
  
\author{Arpad Rimmel}
\orcid{0009-0009-7028-3644}
\email{arpad.rimmel@lisn.fr}
\affiliation{%
  \institution{Universit\'e Paris-Saclay, CNRS, CentraleSup\'elec, Laboratoire Interdisciplinaire des Sciences du Numérique}
  \city{Gif-sur-Yvette}
  \country{France}
  \postcode{91190}
  }

\author{Bich-Li\^en Doan}
\orcid{0009-0001-5853-2486}
\email{bich-lien.doan@lisn.fr}
\affiliation{%
  \institution{Universit\'e Paris-Saclay, CNRS, CentraleSup\'elec, Laboratoire Interdisciplinaire des Sciences du Numérique}
  \city{Gif-sur-Yvette}
  \country{France}
  \postcode{91190}
  }

\renewcommand{\shortauthors}{Hugo Thimonier, Fabrice Popineau, Arpad Rimmel, and Bich-Li\^en Doan}
\begin{abstract}
Deep learning for tabular data has garnered increasing attention in recent years, yet employing deep models for structured data remains challenging. While these models excel with unstructured data, their efficacy with structured data has been limited. Recent research has introduced retrieval-augmented models to address this gap, demonstrating promising results in supervised tasks such as classification and regression. In this work, we investigate using retrieval-augmented models for anomaly detection on tabular data.
We propose a reconstruction-based approach in which a transformer model learns to reconstruct masked features of \textit{normal} samples. We test the effectiveness of KNN-based and attention-based modules to select relevant samples to help in the reconstruction process of the target sample. Our experiments on a benchmark of 31 tabular datasets reveal that augmenting this reconstruction-based anomaly detection (AD) method with sample-sample dependencies via retrieval modules significantly boosts performance. The present work supports the idea that retrieval module are useful to augment any deep AD method to enhance anomaly detection on tabular data. Our code to reproduce the experiments is made available on \href{https://github.com/hugothimonier/Retrieval-Augmented-Deep-Anomaly-Detection-for-Tabular-Data}{\textcolor{blue!70!black}{GitHub}}.
\end{abstract}


\keywords{Anomaly Detection; Tabular Data; Deep Learning}

\ccsdesc[100]{Security and privacy~Intrusion/anomaly detection and malware mitigation}
\ccsdesc[500]{Computing methodologies~Semi-supervised learning settings}
\ccsdesc[300]{Computing methodologies~Neural networks}


\maketitle

\section{Introduction}
    \label{introduction}

        Semi-supervised anomaly detection (AD) consists in learning to characterize a \textit{normal} distribution using a dataset only composed of samples belonging to the \textit{normal}\footnote{The term normal relates to the concept of normality, in opposition to anomaly.} class, in order to identify in a separate dataset the samples that do not belong to this \textit{normal} distribution, namely anomalies. This class of algorithms is often used when the imbalance between classes is too severe, causing standard supervised approaches to fail \cite{Yanminsun2011}. Examples of such applications are cyber intrusion detection \cite{cyber_intrusion}, fraud detection on credit card payment \cite{hilalfraud, thimonier2023comparative}, or tumor detection on images \cite{health_ad}. On the contrary, unsupervised anomaly detection refers to identifying anomalies in a dataset without using labeled training data. These algorithms aim to discover patterns or structures in the data and flag instances that deviate significantly from these patterns. The application of such an approach usually includes detecting mislabeled samples or removing outliers from a dataset that may hinder a model's training process.
        
        While deep learning methods have become ubiquitous and are widely used in the industry for various tasks on unstructured data, relying on deep models for tabular data remains challenging. Indeed, \citet{grinsztajn2022why} discuss how the inherent characteristics of tabular data make this type of data challenging to handle by standard deep models. Hence, recent research on deep learning for structured data \cite{Shavitt2018, tabnet2019, saint2021, kossen2021self, gorishniy2023tabr} has been oriented towards proposing novel training frameworks, regularization techniques or architectures tailored for tabular data. Similarly, general AD methods appear to struggle with tabular data, while the best-performing AD algorithms on tabular data involve accounting for the particular structure of this data type. For instance, \cite{goad, neutralad, shenkar2022anomaly, thimonier2024beyond} put forward self-supervised anomaly detection algorithms targeted for tabular data that significantly outperform general methods on most tested datasets.
        
        In particular, recent research has emphasized the pivotal role of combining feature-feature and samples-sample dependencies in fostering deep learning model's performance on tabular data \cite{gorishniy2023tabr, kossen2021self, saint2021}.
        Following these recent findings, we investigate the benefits of including \textbf{external retrieval modules} to leverage sample-sample dependencies to augment existing AD methods. External retrieval modules can be considered instrumental as they \textbf{can augment any existing model} that may only rely on feature-feature dependencies. In contrast, models that rely on internal retrieval mechanisms are bound to some inductive biases and cannot be used to learn all possible tasks that may be relevant for anomaly detection. 

        Leveraging both types of dependencies is critical to detect all types of anomalies effectively and can increase consistency across datasets, as we empirically show in section \ref{sec:theoretical_motiv}. \citet{han2022adbench} categorize anomalies in tabular data into $4$ families of anomalies which require different types of dependencies to be correctly identified. First, \textit{dependency anomalies} explicitly refers to samples that do not follow the dependency structure that \textit{normal} data follow require feature-feature dependencies to be efficiently identified.
        Second, \textit{global anomalies} refer to unusual data points that deviate significantly from the norm. Relying on both dependencies should improve a model's capacity to detect these anomalies.
        Third, \textit{local anomalies} that refer to the anomalies that are deviant from their local neighborhood can only be identified by relying on sample-sample dependencies. 
        Finally, \textit{clustered anomalies}, also known as group anomalies, are composed of anomalies that exhibit similar characteristics. This type of anomaly requires feature-feature dependencies to be identified. 

        We test the relevance of external modules by employing transformers \cite{attentionisallyouneed} in a mask-reconstruction framework to construct an anomaly score as it was proven to offer strong anomaly detection performance \cite{thimonier2024beyond}. We implement several external retrieval methods to augment the vanilla transformer and evaluate the performance of each approach on an extensive benchmark of tabular datasets.

        We empirically show that the tested approaches incorporating retrieval modules to account for the sample-sample relations outperform the vanilla transformer that only attends to feature-feature dependencies. Furthermore, we propose an empirical experiment to account for the pertinence of combining dependencies, showing that detecting some types of anomalies can require a particular type of dependency.

        The present work offers the following contributions:
        \begin{itemize}
        \itemsep0em 
            \item[•] We propose an extensive evaluation of retrieval-based methods for AD on tabular data.
            \item[•] We empirically show that augmenting existing AD methods with a retrieval module to leverage sample-sample dependencies can help improve detection performance.
            \item[•] We compare our approach to existing methods found in the literature and observe that our method obtains competitive performance metrics.
            \item[•] We provide an explanation as to why combining dependencies leads to better identification of anomalies in tabular data.
        \end{itemize}

    \section{Related works}
    \label{related-works}

        \citet{ruffUnifyingReviewDeep2021a} discuss how anomaly detection bears several denominations that more or less designate the same class of algorithms: anomaly detection, novelty detection, and outlier detection. The literature comprises $4$ main classes of anomaly detection algorithms: density estimation, one-class classification, reconstruction-based, and self-supervised algorithms.

        \paragraph{Density estimation}
        It is often seen as the most direct approach to detecting anomalies in a dataset. The density estimation approach consists in estimating the \textit{normal} distribution and flagging low probability samples under this distribution as an anomaly. Existing methods include using Copula as the COPOD method proposed in \cite{Li2020}, Local Outlier Factor (LOF) \cite{lof}, Energy-based models \cite{ebm_ad} flow-based models \cite{Liu_Tan_Zhou_2022}.

        \paragraph{Reconstruction-based methods}
        Other anomaly detection methods focus on learning to reconstruct samples belonging to the \textit{normal} distribution. In inference, the capacity of the model to reconstruct an unseen sample is used as a measure of anomalousness. The more capable a model is to reconstruct a sample, the more likely the sample is to belong to the \textit{normal} distribution seen in training. Such approach include methods involving autoencoders \cite{chen2018unsupervised, Kim2020RaPP}, diffusion models \cite{diff_medical, zhang2023diffusionad}, GANs \cite{anogan} or attention-based models \cite{thimonier2024beyond}.

        \paragraph{One-Class Classification} One-class classification describes the task of identifying anomalies without \textit{directly} estimating the \textit{normal} density. This class of algorithm involves discriminative models that directly estimate a decision boundary. In \cite{ocsvm, svdd, deep-svdd}, the authors propose algorithms that estimate the support of the \textit{normal} distribution, either in the original data space or in a latent space. During inference, one flags the samples outside the estimated support as anomalies. Other one-class classification methods include tree-based approaches such as isolation forest (IForest) \citep{isolationforest}, extended isolation forest \citep{ief}, RRCF \citep{Guha2016} and PIDForest \citep{Gopalan2019PIDForestAD}. Other methods include approaches to augment existing one-class classification methods with a classifier by generating synthetic anomalies during training, such as DROCC \cite{goyalDROCCDeepRobust2020}.

        \paragraph{Self-Supervised Approaches} Given the recent successes of self-supervision for many tasks, researchers have also investigated using self-supervised methods for anomaly detection. \cite{goad} and \cite{neutralad} propose transformation based anomaly detection methods for tabular data. The former relies on a classifier's capacity to identify which transformation was applied to a sample to measure anomalousness, while the latter relies on a contrastive approach. Similarly, \cite{shenkar2022anomaly} also proposes a contrastive approach to flag anomalies by learning feature-feature relation for \textit{normal} samples. Parallel to this line of work, \cite{sohn2021learning, reiss2021mean} have focused on proposing self-supervised approaches for representation learning tailored for anomaly detection.

        \paragraph{Retrieval modules}
        Retrieval modules have gained attention in recent years in many fields of machine learning. For instance, \cite{retrieval_diffusion} introduces a retrieval module to foster the scalability and efficiency of diffusion models. Parallel to that, \cite{retrieval_cross_task} introduced retrieval for cross-task generalization of large language models, and \cite{retrieval_prompt} introduced it to enhance prompt learning. Finally, retrieval methods have been increasingly used to increase the performance of deep models for tabular data. For instance, \cite{kossen2021self} and \cite{saint2021} introduced internal retrieval modules in deep architecture for supervised tasks on tabular data, while \cite{thimonier2024beyond} relied on internal retrieval modules for anomaly detection. Finally, \cite{gorishniy2023tabr} investigated using external retrieval modules to augment an MLP for supervised tasks on tabular data.
        
        \begin{figure*}
        \begin{center}
        \begin{tikzpicture}

            \node[circle, draw, inner sep=5pt] (z) at (0.5,0) {$\mathbf{z}$};
            \node[align=center] () at (0.5, 0.95) {Masked \\ test sample};
        
            \draw[dashed, rounded corners=8pt, fill=blue!20, fill opacity=0.2] (-0.5,-4.4) rectangle (11,-1.6);
            \node[circle, draw, inner sep=4pt] (sum_circle) at (8.5,-3) {$\Sigma$};
            \node[circle, draw, inner sep=1pt] (add_circle) at (9.75,0) {$+$};
            \node[circle, draw, inner sep=4pt] (e_z) at (4.75,0) {$\mathbf{h}_{\mathbf{z}}$};
            \node[circle, draw, inner sep=4pt] (tilde_z) at (12.5,0) {$\tilde{\mathbf{z}}$};
            \node[align=center] () at (12.5, -0.95) {Reconst. \\ test sample};

            \node[anchor=west] (column) at (0,-3) {$
                \begin{bmatrix}
                \mathbf{x}_1 \\
                \mathbf{x}_2 \\
                \vdots \\
                \mathbf{x}_n
                \end{bmatrix}$ 
            };
                
            \node[anchor=west] (column_encoded) at (4.25, -3){$
                \begin{bmatrix}
                \mathbf{h}_{\mathbf{x}_1} \\
                \mathbf{h}_{\mathbf{x}_2} \\
                \vdots \\
                \mathbf{h}_{\mathbf{x}_n}
                \end{bmatrix}$
                };
            
            \node[rotate=90] () at (-0.25,-3) {Candidate samples};
            
            \node[rectangle, draw, fill=purple, minimum width=1.cm, minimum height=2.2cm, align=center, rounded corners] (embedding_z) at (2,0) {};
            \node[rotate=90] () at (2, 0) {In-Embedding};
            
            \node[rectangle, draw, fill=purple, minimum width=1.cm, minimum height=2.2cm, align=center, rounded corners] (embedding_k) at (2,-3) {};
            \node[rotate=90] () at (2, -3) {In-Embedding};
            
            \node[] () at (2.75, -1.5) {\tiny{\textbf{Shared weights}}};
            
            \node[rectangle, draw, fill=lime, minimum width=1.cm, minimum height=1.cm, align=center, rounded corners] (encoder_z) at (3.5,0) {$E$};
            \node[align=center] () at (3.5, 1) {Transformer \\ Encoder};
        
            \node[rectangle, draw, fill=lime, minimum width=1.cm, minimum height=1.cm, align=center, rounded corners] (encoder_x) at (3.5,-3) {$E$};
            
            \node[rectangle, draw, fill=pink, minimum width=1.cm, minimum height=2.5cm, align=center, rounded corners] (decoder) at (11.,0) {};
            \node[rotate=90] () at (11., 0) {Out-Embedding};
            
            \node[rectangle, draw, fill=lavande, minimum width=1.25cm, minimum height=2.25cm, align=center, rounded corners] (retrieval) at (6.75,-3) {Retrieval \\ Module};
        
            \node[align=center,] () at (9.7,-4) {External retrieval \\module};
            
            \draw[->] (e_z.east) to[out=0, in=90] (retrieval.north);
            \draw[->] (z.east) -- (embedding_z.west);
            \draw[->] ([xshift=-1.5mm]column.east) -- (embedding_k.west);
            \draw[->] (embedding_k.east) -- (encoder_x.west);
            \draw[->] (embedding_z.east) -- (encoder_z.west);
            \draw[dotted] ([xshift=1mm]embedding_z.south) -- ([xshift=1mm]embedding_k.north);
            \draw[dotted] ([xshift=-1mm]embedding_z.south) -- ([xshift=-1mm]embedding_k.north);
            \draw[dotted] ([xshift=1mm]encoder_z.south) -- ([xshift=1mm]encoder_x.north);
            \draw[dotted] ([xshift=-1mm]encoder_z.south) -- ([xshift=-1mm]encoder_x.north);
            \draw[->] (encoder_x.east) -- ([xshift=1.5mm]column_encoded.west);
            \draw[->] ([xshift=-1.5mm]column_encoded.east) -- (retrieval.west);
            \draw[->] (retrieval.east) -- (sum_circle.west);
            \draw[->] (e_z.east) -- (add_circle.west) node[midway, above] {$(1-\lambda)$};
            \draw[->] (sum_circle.east) to[out=0, in=270] node[left] {$\lambda$} (add_circle.south);
            \draw[->] (encoder_z.east) -- (e_z.west);
            \draw[->] (add_circle.east) -- (decoder.west);
            \draw[->] (decoder.east) -- (tilde_z.west);
            
            
            \end{tikzpicture}
            \caption{Forward pass for sample $\mathbf{z}$, see section \ref{subsec:training_procedure} for more detail on training procedure. In the case of no retrieval module, the prediction for a sample $\mathbf{z}$ consists of the upper part of the figure with $\lambda=0$. }
            
            \label{fig:inference}
            \end{center}
        \end{figure*}
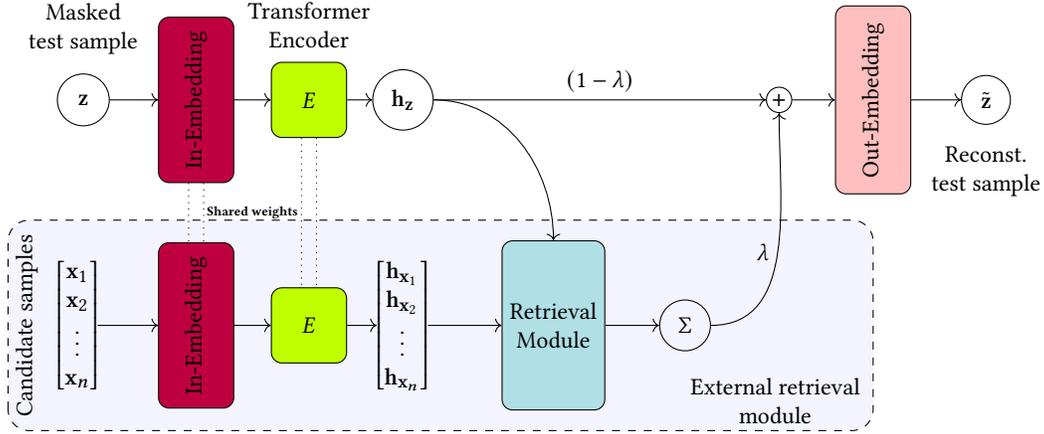

\section{Method}
\label{sec:method}

    \subsection{Learning Objective}
    \label{subsec:learning_objective}

        Let $\mathcal{D}_{train} = \{\mathbf{x}_i \in \mathbb{R}^{d}\}_{i=1}^n$ represent the training set composed of $n$ \textit{normal} samples with $d$ features. The standard approach to anomaly detection involves learning some function $\phi_\theta:\mathbb{R}^d\to \mathcal{Z}$ by minimizing a loss function $\mathcal{L}$. The chosen loss function and the space $\mathcal{Z}$ will vary according to the class of the considered anomaly detection algorithm. Nevertheless, the overall aim of $\mathcal{L}$ is to characterize the distribution of the samples in the training set as precisely as possible. Depending on the chosen AD algorithm, the obtained representation $\phi_\theta(\mathbf{x})$ of sample $\mathbf{x}$ can be used directly or indirectly to compute an anomaly score. 
        
        Formally, the training objective can be summarized as follows
        \begin{equation}
            \label{eq:ad_objective}
            \min_{\theta \in \Theta} \sum_{\mathbf{x} \in \mathcal{D}_{train}} \mathcal{L}(\mathbf{x}, \phi_\theta(\mathbf{x})),
        \end{equation}
        where $\mathcal{L}(\mathbf{x}, \phi_\theta(\mathbf{x}))$ will vary according to the chosen task. For a reconstruction-based method, $\mathcal{Z}$ can be the original data space and $\mathcal{L}$ can be the squared $\ell_2$-norm of the difference between the original sample $\mathbf{x}$ and its reconstructed counterpart, $\mathcal{L}(\mathbf{x}, \phi_\theta(\mathbf{x}))=\|\mathbf{x} - \phi_\theta(\mathbf{x})\|^2$.
        
        Introducing an external retrieval module permits keeping the original objective unchanged while augmenting $\phi_\theta$ with sample-sample dependencies through non-parametric mechanisms. The model involves non-parametric relations as it leverages the entire training dataset to make its prediction. Hence, the model can conjointly attend to feature-feature and sample-sample interactions to optimize its objective.
        
        Formally, instead of minimizing the loss as described in eq. \eqref{eq:ad_objective}, the parameters $\theta$ of the function $\phi_\theta$ are optimized to minimize the loss function as follows
        \begin{equation}
            \label{eq:non_param_ad_objective}
            \min_{\theta \in \Theta} \sum_{\mathbf{x} \in \mathcal{D}_{train}} \mathcal{L}\left(\mathbf{x}, \phi_\theta\left(\mathbf{x}; \mathcal{D}_{train}\right)\right).
        \end{equation}
    
        Nevertheless, not all approaches to AD may benefit from such non-parametric mechanisms. Some pretext tasks involving sample-sample dependencies may lead to degenerate solutions, \textit{e.g.} approaches based on contrastive learning such as the approaches of \cite{neutralad} or \cite{shenkar2022anomaly}. However, reconstruction-based AD methods appear as a natural class of algorithms that may benefit from these non-parametric mechanisms.

    \paragraph{Mask Reconstruction}
        In the mask reconstruction context, we empirically investigate the pertinence of external retrieval modules, as detailed in section \ref{subsec:ret}. Our approach includes stochastic masking which consists in masking each entry in a sample vector $\mathbf{x} \in \mathbb{R}^d$ with probability $p_{mask}$ while setting as the objective task the prediction of the masked-out features from the unmasked features. Formally, we sample $\mathbf{m}\in \mathbb{R}^{d}$ a binary mask vector taking value $1$ when the corresponding entry in $\mathbf{x}$ is masked, $0$ otherwise. This mask $\mathbf{m}$ is then used to construct $\mathbf{x}^m, \mathbf{x}^o \in \mathbb{R}^d$ representing respectively the masked and unmasked entries of sample $\mathbf{x}$. $\mathbf{x}^m, \mathbf{x}^o$ are obtained as follows,
        \begin{equation}
            \begin{array}{rl}
                 \mathbf{x}^m & = \mathbf{m} \odot \mathbf{x}  \\
                 \mathbf{x}^o  & = (\mathbf{1}_d-\mathbf{m}) \odot \mathbf{x},
            \end{array}
        \end{equation}
        where $\mathbf{1}_d$ is the $d$-dimensional unit vector.
                \begin{figure*}[t!]
            \centering
            
              \subfigure[F1-score ($\uparrow$)]{
                \includegraphics[width=0.48\linewidth]{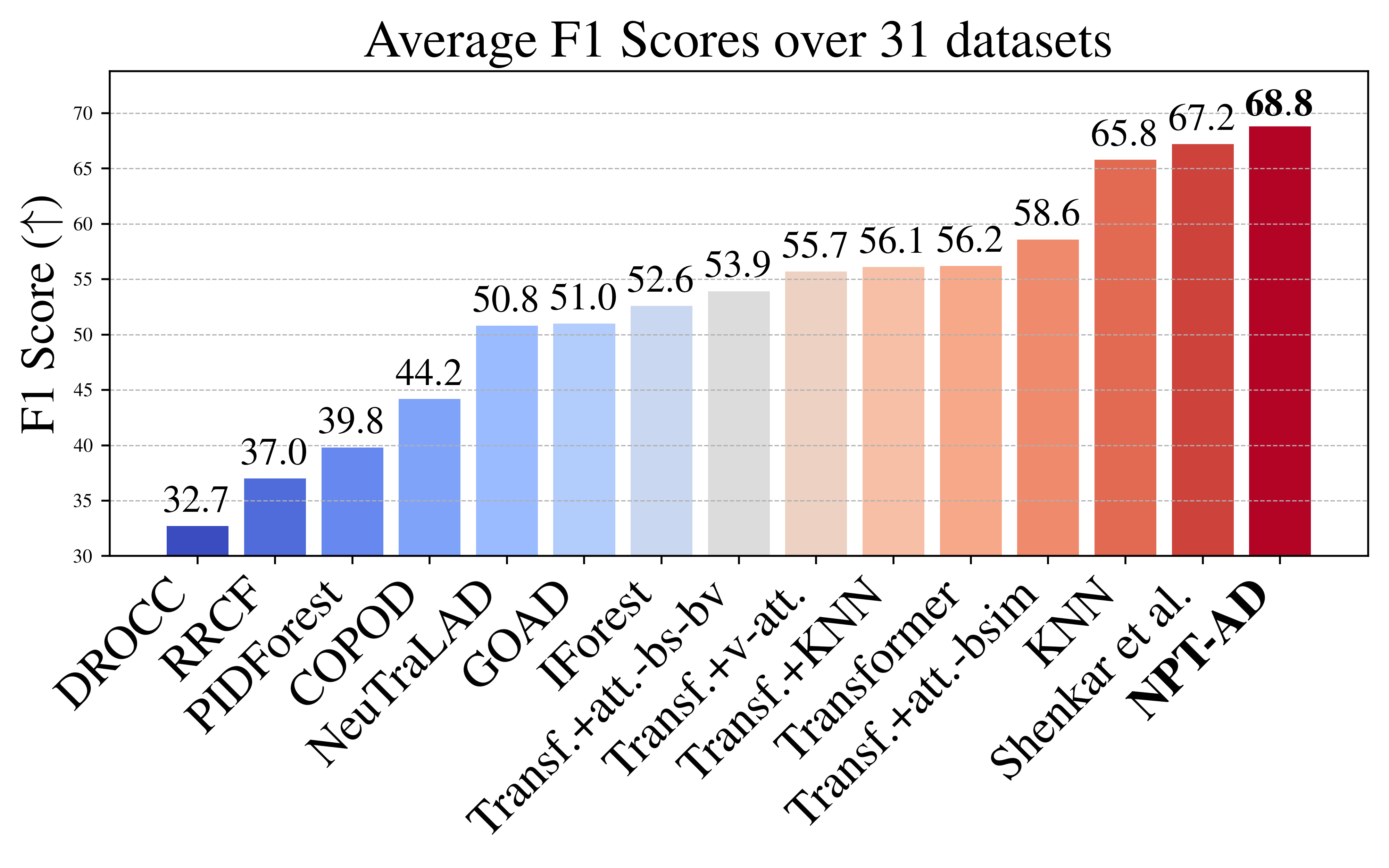}
                \label{fig:sub_avg_F1}
              }
              \hfill
              \subfigure[Rank ($\downarrow$)]{
                \includegraphics[width=0.48\linewidth]{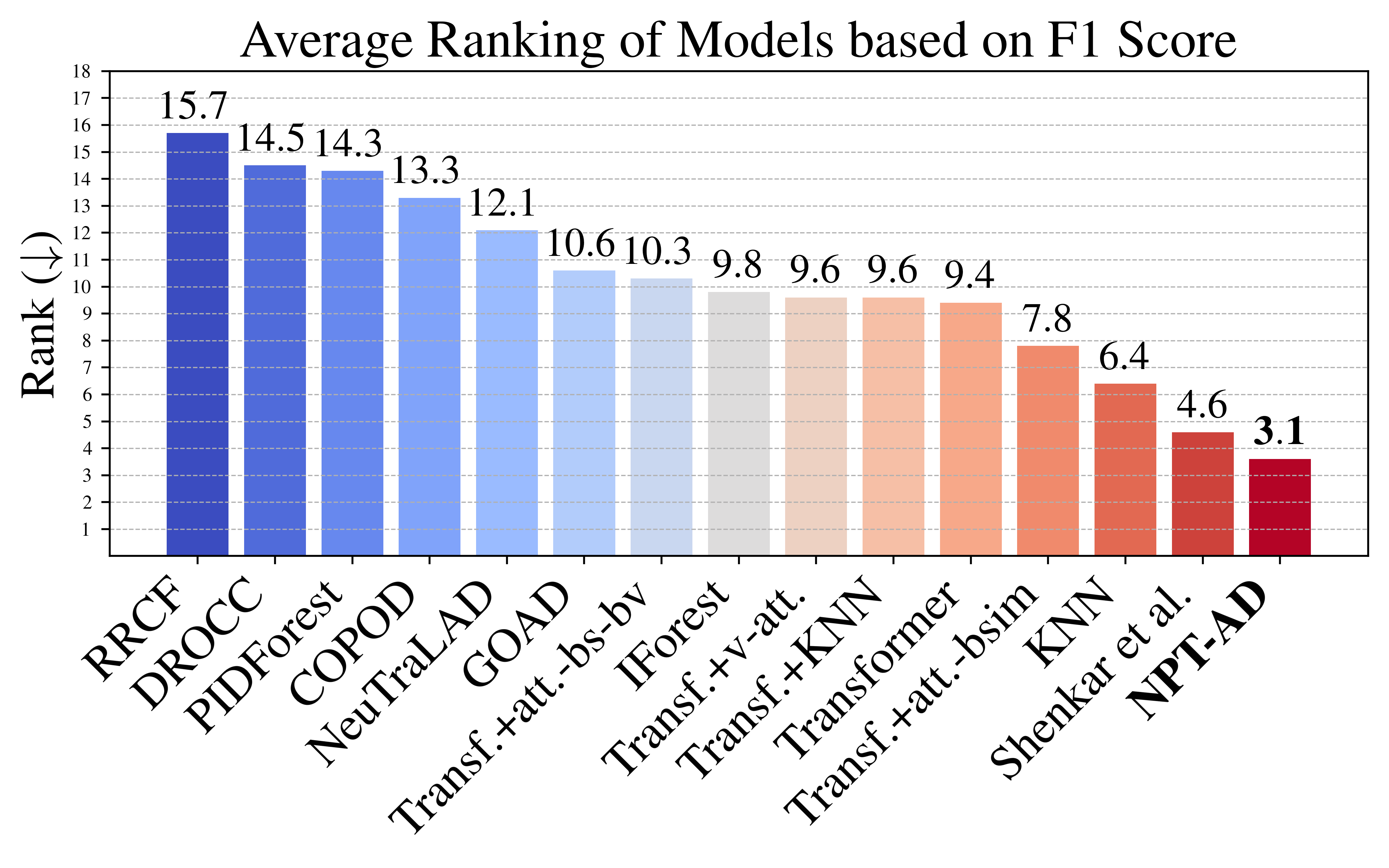}
                \label{fig:sub_avg_rk}
            }
              \caption{For each of the $31$ datasets on which models were evaluated, we report the average F1-score over $20$ runs for $20$ different seeds. We refer readers to \citet{thimonier2024beyond} for details on the obtained metrics and the hyperparameters used for each method. For both figures, the model displayed on the far left is the worst-performing model for the chosen metric, and the one on the far right is the best-performing model. We also highlight the metric of the best-performing model in bold.}
              \label{fig:bench_results}
    \end{figure*}
        
        A model $\phi_\theta : \mathbb{R}^d \times \{0,1\}^d \to \mathbb{R}^d$ is trained to reconstruct the mask features $\mathbf{x}^m$ from its unmasked counterpart $\mathbf{x}^o$ and the mask vector $\mathbf{m}$. By construction, $\phi_\theta(\mathbf{x}^o; \mathbf{m})$ only has non-zero values for corresponding masked features in $\mathbf{m}$. 
         
        In the present work, we evaluate the benefit of replacing the traditional reconstruction learning objective defined as
        \begin{equation}
            \label{eq:masked_recon}
            \min_{\theta \in \Theta} \sum_{\mathbf{x} \in \mathcal{D}_{train}} d(\mathbf{x}^m, \phi_\theta(\mathbf{x}^o; \mathbf{m})),
        \end{equation}
        where $d(.,.)$ is a distance measure;
        with its equivalent augmented with a retrieval module as follows
        \begin{equation}
            \label{eq:masked_recon_nptad}
            \min_{\theta \in \Theta} \sum_{\mathbf{x} \in \mathcal{D}_{train}} d\left(\mathbf{x}^m , \phi_\theta\left( \mathbf{x}^o; \mathbf{m}, \mathcal{D}_{train}^o\right) \right),
        \end{equation}
        where $\mathcal{D}_{train}^o = \{\mathbf{x}_i^o \in \mathbb{R}^d \}_{i=1}^n$. In inference, $\mathcal{D}_{train}^o$ is replaced by $\mathcal{D}_{train}$ in which none of the features of the training samples are masked.

   \begin{table*}[h!]
        \centering
        \caption{Comparison of transformer-based methods. We observe that the external retrieval module $\texttt{attention-bsim}$ significantly improves the AD performance of the vanilla transformer by $4.3\%$ regarding the F1-Score and $1.2\%$ for the AUROC.}
        \label{tab:rez_ret}
        \begin{tabular}{cccccc}
            \toprule
                  & Transformer & $+\texttt{KNN}$ & $+\texttt{v-att.}$ & $+\texttt{att-bsim}$ &
                 $+\texttt{att-bsim-bval}$\\
            \midrule
                 F1-Score ($\uparrow$)
                 & $56.2$
                 & $56.1$
                 & $55.7$
                 & $\mathbf{58.6}$
                 & $53.9$
                 \\ 
                 AUROC ($\uparrow$)
                 & $83.4$
                 & $83.1$
                 & $83.1$
                 & $\mathbf{84.4}$
                 & $82.1$
                 \\
            \bottomrule
        \end{tabular}
    \end{table*}

    \subsection{Retrieval methods}
    \label{subsec:ret}

        Let $\mathbf{z}$ denote the sample of interest for which we wish to reconstruct its masked features $\mathbf{z}^m$ given its observed counterpart $\mathbf{z}^o$. Let $\mathcal{C}$ denote the candidate samples from the training set from which $k$ \textit{helpers} are to be retrieved, and $\mathcal{H}$ the retrieved \textit{helpers}, $\mathcal{H} \subseteq \mathcal{C}$. 
    
        We consider several \textit{external} retrieval modules that rely on similarity measures to identify relevant samples to augment the encoded representation of the sample of interest $\mathbf{z}$. It involves placing a retrieval module after the transformer encoder and before the output layer, as shown in figure \ref{fig:inference}. We investigate in section \ref{sec:discussion} the impact of modifying the location of the retrieval module and consider placing it before the encoder as an alternative. 
        \\ For each method, the retrieval module consists in selecting the top-$k$ elements that maximize a similarity measure $\mathcal{S}(\cdot,\cdot)$ and use a value function to obtain representations of the chosen samples $\mathcal{V}(\cdot,\cdot)$ to be aggregated with sample $\mathbf{z}$.
    
        \paragraph{KNN-based module}
        First, we consider a simple method that identifies the $k$ most relevant samples in $\mathcal{C}$ using a $\texttt{KNN}$ approach. Formally, the similarity and value functions are defined as follows
        \begin{equation}
        \label{eq:knn}
        \begin{aligned}
            \mathcal{S}(\mathbf{z}, \mathbf{x}) &= -\|\mathbf{h_z} - \mathbf{h_x}\| \\  \mathcal{V}(\mathbf{z}, \mathbf{x}) &= \mathbf{h_x},
        \end{aligned}
        \end{equation}
        where $\mathbf{h_x}$ and $\mathbf{h_z}$ denote the representations of respectively sample $\mathbf{x}$ and $\mathbf{z}$ and $\|.\|$ is the $\ell_2$-norm.
    
        \paragraph{Attention-based modules}
        Second, we consider attention mechanisms to select $\mathcal{H}$. We consider three types of attention inspired by those proposed in \cite{gorishniy2023tabr}. First, the vanilla attention (later referred to as $\texttt{v-attention}$), where the score and value function used to select the retrieved samples are defined as
        \begin{equation}
        \label{eq:v-attention}
        \begin{aligned}
            \mathcal{S}(\mathbf{z}, \mathbf{x}) &= W_Q(\mathbf{h_z})^\top W_K (\mathbf{h}_x) \\  \mathcal{V}(\mathbf{z}, \mathbf{x}) &= W_V(\mathbf{h_x}).
        \end{aligned}
        \end{equation}
        where $W_Q, W_K$ and $W_V$ are learned parameters.\\
        Second, we also consider another type of attention module, later referred to as  $\texttt{attention-bsim}$, which involves replacing the score function defined in eq. \eqref{eq:v-attention} as follows
        \begin{equation}
        \label{eq:attention-bsim}
        \begin{aligned}
            \mathcal{S}(\mathbf{z}, \mathbf{x}) &= -\left \|W_K (\mathbf{h_z}) - W_K (\mathbf{h_x})\right \|^2\\ 
            \mathcal{V}(\mathbf{z}, \mathbf{x}) &= W_V(\mathbf{h_x}).
        \end{aligned}
        \end{equation}
        Third, we consider $\texttt{attention-bsim-bval}$ a modification of the value function in eq. \eqref{eq:attention-bsim} as
        \begin{equation}
        \label{eq:attention-bsim-bval}
        \begin{aligned}
            \mathcal{S}(\mathbf{z}, \mathbf{x}) &= -\left \|W_K (\mathbf{h_z}) - W_K (\mathbf{h_x})\right \|^2 \\
            \mathcal{V}(\mathbf{z}, \mathbf{x}) &= T(W_K (\mathbf{h_z}) - W_K (\mathbf{h_x})),
        \end{aligned}
        \end{equation}
        where $T(\cdot)= \texttt{LinearWithtoutBias} \circ \texttt{Dropout} \circ \texttt{ReLU} \circ \texttt{Linear} (\cdot) $.
        
        \paragraph{Aggregation} The retrieval modules necessitate aggregating the obtained retrieved representations $\mathcal{V}(\mathbf{z},\mathbf{x})$ with the representation of the sample of interest $\mathbf{z}$. We aggregate the value of the selected top-$k$ \textit{helpers}, to be fed to the final layer
        \begin{equation}
        \label{eq:aggregation}
            \tilde{\mathbf{h}}_\mathbf{z} = (1 - \lambda)\cdot\mathbf{h_z} + \lambda \cdot \frac{1}{k}\sum_{x \in \mathcal{H}} \mathcal{V}(\mathbf{z}, \mathbf{x}).
        \end{equation}
        where $\lambda \in [0,1)$ is a hyperparameter.

    \subsection{Anomaly score}
    \label{subsec:anomalyscore}

        We construct an anomaly score to assess whether a test sample belongs to the \textit{normal} distribution or should be considered an anomaly. As a reconstruction-based method, our anomaly score is directly obtained from the optimized loss during training: the better the trained model reconstructs a sample, the more likely the sample is to be \textit{normal}. Indeed, since the model has exclusively seen \textit{normal} samples during training, it should be less able to reconstruct anomalies correctly since they stem from a different distribution. On the contrary, unseen \textit{normal} samples should be well reconstructed. We rely on the squared $\ell_2$-norm of the difference between the reconstructed sample and the original sample for numerical features, while we use the cross-entropy loss function for categorical features.

        We rely on a mask bank composed of $m$ $d$-dimensional masks to construct the anomaly score. We apply each mask to each validation sample and reconstruct the masked features to compute the reconstruction error for each mask. Thus, each validation sample is masked and reconstructed $m$ times. The anomaly score is constructed as the average reconstruction error over the $m$ masks.
        To construct the mask bank, we fix the maximum of features to be masked simultaneously $r$ and and construct $m=\sum_{k=1}^r \binom{d}{k}$ masks.
        Choosing deterministic masks instead of random masks to create the mask bank used for inference is beneficial for two reasons. First, since the model will reconstruct all features at least once, it increases the likelihood of identifying different types of anomalies. Indeed, anomalies that deviate from the \textit{normal} distribution due to a single feature would only be identified if the corresponding mask hiding this feature would be included. Second, this approach ensures that all samples are masked identically to build the anomaly score. We investigate the impact of constructing a random mask bank instead of a deterministic mask bank in section \ref{subsec:random_mask}. 
        \\ We use the whole unmasked training set\footnote{For large datasets, we resort to a random subsample of the training set for computational reasons.} $\mathcal{C}=\mathcal{D}^{train}$ to predict the masked features of each sample for each of the $m$ masked vectors and construct the anomaly score for a validation sample $\mathbf{z}$ as
        \begin{equation}
            \label{eq:npt-ad}
            \mbox{AD-Score}(\mathbf{z}; \mathcal{D}^{train}) = \frac{1}{m} \sum_{k=1}^m \mathcal{L}(\mathbf{z}^{(k)};\mathcal{D}^{train}), 
        \end{equation}
        where $\mathcal{L}(\mathbf{z}^{(k)};\mathcal{D}^{train})$ designates the loss for the sample $\mathbf{z}$ with mask $k$.

    \subsection{Training pipeline}
    \label{subsec:training_procedure}

    Let $\mathbf{x} \in \mathcal{X} \subseteq \mathbb{R}^d$ be a sample with $d$ features, which can be either numerical or categorical. Let $e$ designate the hidden dimension of the transformer. The training pipeline consists of the following steps:
    \paragraph{Masking}
        We sample from a Bernoulli distribution with probability $p_{mask}$ whether each of the $d$ features is masked.
        $$
        \mbox{mask} = (m_1, \dots, m_{d}), 
        $$
        where $m_j \sim \mathcal{B}(1,p_{mask})$ $\forall j \in [1,...,d]$ and $m_j=1$ if feature $j$ is masked.
        
     \paragraph{Encoding}
        For numerical features, we normalize to obtain $0$ mean and unit variance, while we use one-hot encoding for categorical features. At this point, each feature $j$ for $j \in [1,2,...,d]$ has an $e_j$-dimensional representation, $encoded(\mathbf{x}_j) \in \mathbb{R}^{e_j}$ , where $e_j=1$ for numerical features and for categorical features $e_j$ corresponds to its cardinality. We then mask each feature according to the sampled mask vector and concatenate each feature representation with the corresponding mask indicator function. Hence, each feature $j$ has an $(e_j+1)$-dimensional representation 
        $$
        ((1 - m_j) \cdot encoded(\mathbf{x}_j), m_j) \in \mathbb{R}^{e_j+1},
        $$
        where $\mathbf{x}_j$ is the $j$-th features of sample $\mathbf{x}$.
        
    \paragraph{In-Embedding}
        We pass each of the features encoded representations of sample $\mathbf{x}$ through learned linear layers $\texttt{Linear}(e_j+1,e)$. We also learn $e$-dimensional index and feature-type embeddings as proposed in \cite{kossen2021self}. Both are added to the embedded representation of sample $\mathbf{x}$. The obtained embedded representation is thus of dimension $d \times e$
        $$
        h_{\mathbf{x}} = (h_{\mathbf{x}}^{1},h_{\mathbf{x}}^{2},\dots,h_{\mathbf{x}}^{d}) \in \mathbb{R}^{d \times e},
        $$
        and $h_{\mathbf{x}}^{j} \in \mathbb{R}^e$ corresponds to the embedded representation of feature $j$ of sample $\mathbf{x}$.
        
    \paragraph{Transformer Encoder}
        The embedded representation obtained from the previous step is then passed through a transformer encoder. The output of the transformer $\mathbf{h}_{\mathbf{x}}$ is of dimension $d \times e$.
        
    \paragraph{Out-Embedding}
        The output of the transformer, $\mathbf{h}_{\mathbf{x}} \in \mathbb{R}^{d\times e}$ is then used to compute an estimation of the original $d$-dimensional vector. To obtain the estimated feature $j$, we take the $j$-th $d$-dimensional representation which is output by the transformer encoder, $h_{\mathbf{x}_j} \in \mathbb{R}^d$, and pass it through a linear layer $\texttt{Linear}(e, e_j)$, where $e_j$ is $1$ for numerical features and the cardinality for categorical features. 
        
    \paragraph{External Retrieval Modules}
        During training, for a batch $\mathcal{B}$ composed of $b$ samples, for each sample $\mathbf{x} \in \mathcal{B}$, the entire batch serves as the candidates $\mathcal{C}$. In inference, a random subsample of the training set is used as $\mathcal{C}$. Both for training and inference, when possible memory-wise, we use as $\mathcal{B}$ and $\mathcal{C}$ the entire training set.\\
        As input, the retrieval module receives a $\mathbb{R}^{d\times e}$ representation for each sample. Operations described in eq. \eqref{eq:knn}, \eqref{eq:v-attention}, \eqref{eq:attention-bsim} and \eqref{eq:attention-bsim-bval} are performed on the flattened representations of samples $\mathbf{x}$, $\mathbf{h}^{\mathbf{x}}_{flatten}\in\mathbb{R}^{d\cdot e}$. After selecting $\mathcal{H}\subseteq \mathcal{C}$, each sample is transformed back to its original dimension to allow aggregation as described in eq. \eqref{eq:aggregation}.

\section{Experiments}
\label{sec:experiments}
    
    \paragraph{Datasets} We experiment on an extensive benchmark of tabular datasets following previous work \cite{shenkar2022anomaly, thimonier2024beyond}. The benchmark is comprised of two datasets widely used in the anomaly detection literature, namely Arrhythmia and Thyroid, a second group of datasets, the "Multi-dimensional point datasets", obtained from the Outlier Detection DataSets (ODDS)\footnote{\url{http://odds.cs.stonybrook.edu/}} containing 28 datasets. We also include three real-world datasets from \cite{han2022adbench}: fraud, campaign, and backdoor. We display each dataset's characteristics in table \ref{tab:characteristics} in appendix \ref{appendix:dataset_char}. 
    
    \begin{table*}[ht!]
        \begin{center}
        \caption{Comparison of the F1-score ($\uparrow$) of transformer+$\texttt{attention-bsim}$ across values of $k$. Here $-1$ stands for $\mathcal{H}=\mathcal{C}$. Some values are N/A either because it is not relevant to compute or when there are not enough samples in the training set for the selected value of $k$.}
        \label{tab:rez_k}
        \begin{tabular}{c ccccccc}
            \toprule 
                 $k$ 
                 & $0$
                 & $5$	
                 & $25$
                 & $50$
                 & $200$
                 & $500$
                 & $-1$
                 \\
            \midrule
                \multicolumn{8}{c}{transformer+$\texttt{attention-bsim}$}
                \\
            \midrule
                 Abalone 
                 & $42.5\pm$\small{$7.8$}
                 & $53.0\pm$\small{$6.4$}
                 & $\mathbf{54.9}\pm$\small{$5.4$}
                 & $\mathbf{55.0}\pm$\small{$5.4$}
                 & $52.0\pm$\small{$5.6$}
                 & $54.0\pm$\small{$6.5$}
                 & $53.0\pm$\small{$5.7$}
                 \\
                 Satellite
                 & $65.6\pm$\small{$3.3$}
                 & $71.5\pm$\small{$2.4$}
                 & $71.3\pm$\small{$1.3$}
                 & $71.2\pm$\small{$1.6$}
                 & $70.8\pm$\small{$1.8$}
                 & $71.2\pm$\small{$1.7$}
                 & $\mathbf{71.9}\pm$\small{$1.5$}
                 \\
                 Lympho
                 & $88.3\pm$\small{$7.6$}
                 & $91.7\pm$\small{$8.3$}	
                 & $\mathbf{93.3}\pm$\small{$8.2$}
                 & $91.7\pm$\small{$8.3$}
                 & N/A
                 & N/A
                 & $90.0\pm$\small{$8.1$}
                 \\
                 Satimage
                 & $89.0\pm$\small{$4.1$}
                 & $88.8\pm$\small{$3.8$}
                 & $88.4\pm$\small{$3.8$}
                 & $89.4\pm$\small{$4.2$}
                 & $88.8\pm$\small{$4.3$}
                 & $89.1\pm$\small{$4.3$}
                 & $\mathbf{93.2}\pm$\small{$1.7$}
                 \\
                 Thyroid
                 & $55.5\pm$\small{$4.8$}
                 & $\mathbf{56.9}\pm$\small{$5.3$}
                 & $55.9\pm$\small{$5.2$}
                 & $56.3\pm$\small{$5.2$}
                 & $56.4\pm$\small{$5.2$}		
                 & $55.9\pm$\small{$5.6$}
                 & $55.8\pm$\small{$6.3$}
                 \\
                 Cardio
                 & $81.0\pm$\small{$4.1$}
                 & $81.2\pm$\small{$1.6$}
                 & $\mathbf{81.9}\pm$\small{$1.4$}
                 & $\mathbf{81.9}\pm$\small{$1.4$}
                 & $\mathbf{81.9}\pm$\small{$1.4$}
                 & $\mathbf{81.9}\pm$\small{$1.4$}
                 & $80.6\pm$\small{$2.4$}
                 \\
                 Ionosphere	
                 & $88.1\pm$\small{$2.8$}
                 & $89.4\pm$\small{$5.0$}
                 & $90.2\pm$\small{$4.5$}
                 & $89.8\pm$\small{$4.3$}
                 & N/A
                 & N/A	
                 & $\mathbf{91.7}\pm$\small{$2.1$}
                 \\
                 \midrule
                 mean
                 & $70.3$
                 & $76.1$
                 & $\mathbf{76.6}$
                 & $76.5$
                 & $70.0$
                 & $70.4$
                 & $\mathbf{76.6}$
                 \\
                 mean std 
                 & $4.9$
                 & $4.7$
                 & $4.3$
                 & $4.3$
                 & $3.7$
                 & $3.9$
                 & $4.0$ \\
            \bottomrule
        \end{tabular}
    \end{center}
    \end{table*}
    \paragraph{Experimental Settings} Following previous work in the AD literature, \cite{zong2018deep, goad}, we construct the training set with a random subsample of the \textit{normal} samples representing 50\% of the \textit{normal} samples, we concatenate the 50\% remaining with the entire set of anomalies to constitute the validation set. Similarly, we fix the decision threshold for the AD score such that the number of predicted anomalies equals the number of existing anomalies.

    To evaluate to which extent sample-sample dependencies are relevant for anomaly detection, we compare models that attend to relations between samples to the vanilla transformer model. We compare the different methods discussed using the F1-score ($\uparrow$) and AUROC ($\uparrow$) metrics following the literature. For each dataset, we report an average over $20$ runs for $20$ different seeds; we display the detailed results for all five transformer-based methods in tables \ref{tab:detailed_table_f1} and \ref{tab:detailed_table_auc} in appendix \ref{appendix:detailed_rez} and report in Table \ref{tab:rez_ret} the average F1 and AUROC.

    We considered three regimes for the transformer dimensions depending on the dataset size. The transformer encoder comprises $2$ or $4$ layers with $4$ attention heads and hidden dimension $e\in\{8,16,32\}$ for smaller to larger datasets. We train the transformer with a mask probability $p_{mask}$ set to $0.25$ or $0.15$ and rely on the LAMB optimizer \cite{lamb} with  $\beta = (0.9, 0.999)$ and also included a Lookahead \cite{lookahead} wrapper with slow update rate $\alpha = 0.5$ and $k = 6$ steps between updates. We also include a dropout regularization with $p=0.1$ for attention weights and hidden layers. We ensure that during training, all samples from a batch are not masked simultaneously so that the retrieval module receives encoded representations of unmasked samples as it will in the inference stage.
    We considered the same transformer architecture and hyperparameters for the same dataset for each considered approach. For external retrieval modules, we chose for simplicity $\lambda=0.5$ for aggregation as detailed in eq. \eqref{eq:aggregation}. We study in section \ref{subsec:lamnbda_choice} the effect of varying the value of $\lambda$ on the model's performance. For the KNN module, we set $k=5$ as the cardinality of $\mathcal{H}$ since KNN-based anomaly detection methods \cite{ramaswamy2000efficient} with $k=5$ have shown strong anomaly detection performance \cite{shenkar2022anomaly}. In contrast, for the attention modules, we set $\mathcal{H}=\mathcal{C}$ and use the attention weights to compute a weighted mean to be aggregated as in eq. \eqref{eq:aggregation}. We further discuss the choice of $k$ in section \ref{subsec:choice_k}. Finally, depending on the dataset, we trained the model until the loss stopped improving for $50$ or $100$ consecutive epochs.
    Each experiment in the present work can be replicated using the code made available on github\footnote{\textcolor{blue!80!black}{\url{https://github.com/hugothimonier/Retrieval-Augmented-Deep-Anomaly-Detection-for-Tabular-Data}}}.

    \paragraph{Results} As reported in Table \ref{tab:rez_ret}, we observe that not all retrieval modules induce a significant boost in anomaly detection performance. We observe that only the transformer augmented by the $\texttt{attention-bsim}$ module performs significantly better than the vanilla transformer. Indeed, augmenting the vanilla transformer with the retrieval module detailed in eq. \eqref{eq:attention-bsim} allows to increase the average F1-score by $4.3$\% and AUROC by $1.2$\%. This result is all the more interesting since it contradicts the results obtained for the supervised classification and regression tasks investigated in previous work \cite{gorishniy2023tabr} where the module that obtains the best performance is $\texttt{attention-bsim-bval}$. However, let us mention that the $\texttt{attention-bsim-bval}$ module involved in our work is not identical to the one put forward in \cite{gorishniy2023tabr} as it does not involve any label. 

    For completeness, we compare the different architectures proposed in the present work to existing methods in the literature. We rely on the experiments conducted in \cite{thimonier2024beyond, shenkar2022anomaly} for the metrics of the competing methods. We display in figure \ref{fig:bench_results} the comparison to existing methods. We compare our methods to recent deep methods, namely GOAD \cite{goad}, DROCC \cite{goyalDROCCDeepRobust2020}, NeuTraL-AD \cite{neutralad}, the contrastive approach proposed in \cite{shenkar2022anomaly} and NPT-AD \cite{thimonier2024beyond}. We also compare to classical non-deep methods such as Isolation Forest \cite{isolationforest}, KNN \cite{ramaswamy2000efficient}, RRCF \cite{Guha2016}, COPOD \cite{Li2020} and PIDForest \cite{Gopalan2019PIDForestAD}. We refer the reader to \cite{thimonier2024beyond} for the detail on the F1-score per dataset for other methods than those shown in Tables \ref{tab:detailed_table_f1} and \ref{tab:detailed_table_auc} in appendix \ref{appendix:detailed_rez}.

\section{Discussion}
\label{sec:discussion}

        \subsection{Why combine dependencies?}
        \label{sec:theoretical_motiv}

            To account for the fact that the retrieval-augmented transformer outperforms the vanilla transformer, we hypothesize that \textit{different types of anomalies require different dependencies to identify them}. In this section, we provide a simple example to demonstrate this statement.
            Consider a simple three dimensional data space, $\mathbf{x} = (x_1, x_2, x_3) \in\mathbb{R}^3$, in which the relation between the features of \textit{normal} sample are defined as follows,
            \begin{equation}
                \label{eq:feature_relation}
                \begin{array}{cc}
                     x_2 = & \alpha_1 + \beta_1  x_1 + \varepsilon \\
                     x_3 = & \alpha_2 + \beta_2  x_2^2 + \varepsilon,
                \end{array}
            \end{equation}
            where $\varepsilon$ is some white noise and $(\alpha_1,\alpha_2, \beta_1,\beta_2)\in\mathbb{R}^2$ are scalars.
          
\begin{figure}[t]
                \begin{center}
                    \includegraphics[scale=0.5]{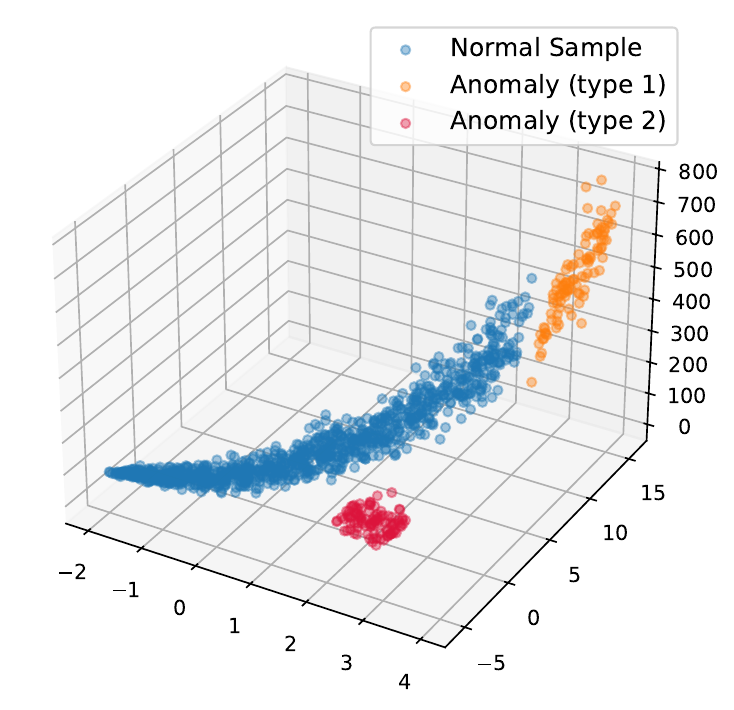}  
                \end{center}
                \caption[Illustration of the pertinence of combining dependencies]{Anomalies of type $1$ (\tikz{\filldraw[orange] (0,0) circle (2pt);}) require inter-sample dependencies to be correctly identified with high probability.
                Anomalies of type $2$ (\tikz{\filldraw[purple] (0,0) circle (2pt);}) on the other hand require inter-feature dependencies to be correctly identified.}
                \label{fig:why_dep}
            \end{figure}

            Let us consider two types of anomalies as shown in Figure \ref{fig:why_dep}. First anomalies of type $1$, in which the relations between features are identical to the ones given in eq. \eqref{eq:feature_relation} but in a different subspace. Now consider type $2$ anomalies, for which the values of the generating feature $x_1$ are in the same subspace as \textit{normal} samples, the relation between $x_1$ and $x_2$ is the same, but the parameters of the relation between $x_2$ and $x_3$ differ. Type $1$ (resp. $2$) anomalies are akin to \textit{local anomalies} (resp. \textit{dependency anomalies}) discussed in \cite{han2022adbench}.

             \begin{table}[t!]
                \begin{center}
                    \caption{Share (\%) of each class correctly identified ($\uparrow$). Average over $5$ data splits. The Table should be read as follows: On average, the transformer correctly classified $78.5\%$ of type $1$ anomalies as anomalies.}
                    \label{tab:theoritical}
                    \begin{tabular}{cccc}
                    \toprule
                          & Normal & Anomalies & Anomalies \\
                          & & (type $1$) & (type $2$) \\
                        \midrule
                        Mask-KNN  & $93.0\%$ \small{$(\pm0.4)$} & $100.0\%$\small{$(\pm0.0)$} & $77.3\%$\small{$(\pm4.4)$} \\
                         Transformer & $91.4\%$ \small{$(\pm1.4)$} & $78.5\%$\small{$(\pm1.0)$} & $100.0\%$\small{$(\pm0.0)$} \\
                         
                         $+\texttt{att-bsim}$& $94.6\%$ \small{$(\pm0.5)$} & $88.0\%$\small{$(\pm0.8)$} & $100.0\%$\small{$(\pm0.0)$} \\
                         \bottomrule
                    \end{tabular}
                \end{center}
            \end{table}

            To test our hypothesis, we compare the retrieval augmented transformer to the vanilla transformer and Mask-KNN, a reconstruction-based technique introduced in \cite{thimonier2024beyond}, that relies on KNN imputation to reconstruct masked features. Mask-KNN (resp. the transformer) can be considered approximately equivalent to the retrieval augmented transformer without considering the feature-feature dependencies (resp. the sample-sample dependencies).

            In the present framework, models only leveraging inter-feature relations, such as the vanilla transformer, may have limited capacities to identify anomalies if they satisfy the same relations as given in eq. \eqref{eq:feature_relation} but in a different subspace, i.e., anomalies of type $1$. 
            Similarly, a model that only relies on inter-sample relations, e.g., Mask-KNN, would struggle to correctly identify anomalies of type $2$ as they lie in a subspace close to normal samples. 
            
            \paragraph{Synthetic Dataset} We effectively test this hypothesis by constructing a synthetic three-dimensional dataset where the features of normal samples satisfy the relations described in eq. \eqref{eq:feature_relation}. We also construct two types of anomalies where type $1$ anomalies follow the same inter-feature relation as normal samples but with values in a non-overlapping interval as those of the normal class. Type $2$ anomalies are constructed to be close to the normal population but with inter-feature relation differing from eq. \eqref{eq:feature_relation}. This synthetic dataset is displayed in figure \ref{fig:why_dep}. The normal population comprises $1000$ samples, and we generate $200$ anomalies for each type. We use half of the normal samples to train models and use the rest merged with the anomalies as the validation set. We compare the capacity of Mask-KNN, the vanilla transformer, and the retrieval-augmented transformer to identify anomalies and display the obtained results in Table \ref{tab:theoritical}. We consider the same mask bank, $\{(1,0,0),(0,1,0),(0,0,1)\}$, for all three approaches. We use the same strategy for selecting the decision threshold as detailed in section \ref{sec:experiments}. See appendix \ref{app:exp_synthetic_dataset} for more details on the experimental setting.
            
             \paragraph{Results} We observe in Table \ref{tab:theoritical} that the vanilla transformer struggles to detect type $1$ anomalies in comparison with other methods, as they explicitly require inter-sample dependencies to be identified. Nevertheless, it correctly identifies $91.4$\% of normal samples on average and perfectly detects type $2$ anomalies. Conversely, Mask-KNN obtains significantly lower performance than competing methods for type $2$ anomalies while correctly identifying type $1$ anomalies and normal samples. Notice how both methods cannot correctly identify anomalies from one of the two anomaly types despite using a perfectly separable dataset. On the contrary, the retrieval augmented transformer can better identify both types of anomalies as it can leverage inter-sample and inter-feature dependencies. This experiment provides information as to why combining dependencies for anomaly detection is relevant: it allows the detection of most anomaly types while other approaches are confined to some anomaly categories.
    
        \subsection{Analysis of the effect of the number of helpers $k$ on the performance}
        \label{subsec:choice_k}

        Note that we randomly selected a subset of the dataset from the benchmark for the ablation studies conducted hereafter. We use this subset of datasets in each experiment for computational reasons and to avoid cherry-picking the hyperparameters.

        We investigate the impact of varying $k$, the number of \textit{helpers}, for both the transformer augmented by $\texttt{attention-bsim}$ and the KNN module since they are the two best-performing retrieval-augmented models. We keep hyperparameters constant and only make $k$ vary between runs. We report in Table \ref{tab:rez_k} the obtained results for both architectures for values of $k\in \{0,5,25,50,200,500,-1\}$, where $-1$ implies that $\mathcal{H}=\mathcal{C}$.

        A noticeable trend is that both architectures obtain the best results with moderate numbers of helper, \textit{i.e.} $k\in\{25,50\}$, and have worse performance for smaller values of $k$. This observation suggests that performance displayed in tables \ref{tab:rez_ret},\ref{tab:detailed_table_f1}, and \ref{tab:detailed_table_auc} 
        could be improved for optimized values of $k$.

    \subsection{Analysis of the effect of the value of $\lambda$}
        \label{subsec:lamnbda_choice}
        
        We analyze the performance variation for the best-performing external retrieval module, namely transformer+$\texttt{attention-bsim}$, for varying values of $\lambda$. We compute the average F1-score over $10$ runs for each value of $\lambda$ in $\{0.1,0.2,\dots,0.7\}$ while keeping the same hyperparameters. We report the results in Table \ref{tab:rez_lambda}.
        
        We observe a slight variation of the average metric across the datasets when setting lambda values from $0.1$ to $0.7$. The maximum value is obtained for $0.5$, but the difference with other values of $\lambda$ is non-significant. Nevertheless, we observe significant differences between the obtained results for isolated datasets for different values of $\lambda$. This observation supports the idea that an optimal value of $\lambda$ exists for each dataset, which may differ between datasets.
    
   \subsection{Location of the retrieval module}
        \label{subsec:loc_ret_module}

    \begin{table}[!t]
    \begin{center}
        \caption{Comparison of the performance of the transformer + $\texttt{attention-bsim}$ model for different retrieval module architecture based on the F1-score ($\uparrow$).}
        \label{tab:ret_loc}
            \begin{tabular}{cccc}
            \toprule
                 Retrieval & $\texttt{post-enc}$ & $\texttt{post-emb}$ & $\texttt{post-emb}$ \\
                 Aggregation & $\texttt{post-enc}$ & $\texttt{post-enc}$ & $\texttt{post-emb}$ \\
            \midrule
                Abalone 
                & $\mathbf{53.0}\pm$\small{$5.7$}
                & $47.0\pm$\small{$7.6$}
                & $30.5\pm$\small{$14$}
                \\
                Satellite 
                & $\mathbf{71.9}\pm$\small{$1.5$}
                & $60.5\pm$\small{$0.9$}
                & $65.1\pm$\small{$2.3$}
                \\
                Lympho
                & $\mathbf{90.0}\pm$\small{$8.1$}
                & $\mathbf{91.6}\pm$\small{$8.3$}
                & $84.2\pm$\small{$11.1$}
                \\
                Satimage
                & $\mathbf{93.2}\pm$\small{$1.7$}
                & $50.8\pm$\small{$17.0$}
                & $65.9\pm$\small{$15.3$}
                \\
                Thyroid
                & $55.8\pm$\small{$6.3$}
                & $55.2\pm$\small{$5.9$}
                & $\mathbf{58.1}\pm$\small{$5.0$}
                \\
                Ionosphere
                & $\mathbf{91.7}\pm$\small{$2.1$}
                & $88.2\pm$\small{$1.4$}
                & $\mathbf{91.3}\pm$\small{$2.0$}
                \\
            \midrule
                mean 
                & $\mathbf{75.9}$
                & $65.6$
                & $65.9$ \\
                mean std
                & $\mathbf{4.2}$
                & $6.9$
                & $8.3$
            \\
            \bottomrule
            \end{tabular}
        \end{center}
        \end{table}
        We investigate the impact of the retrieval module's location on the retrieval-augmented models' performance. To do so, we focus on the transformer+$\texttt{attention-bsim}$ model since it has shown to be the best-performing retrieval-based method. We compare three different architectures: 
        \begin{itemize}
            \item ($\texttt{post-enc, post-enc}$) for \textit{post-encoder} location and \textit{post-encoder} aggregation: the architecture detailed in figure \ref{fig:inference},
            \item ($\texttt{post-emb, post-enc}$), the architecture in which the retrieval module is located after the embedding layer, but the aggregation is still located after the encoder,
            \item ($\texttt{post-emb}$, $\texttt{post-emb}$) the architecture where both retrieval and aggregation are located after the embedding layer.
        \end{itemize}
        
        We do not report the results for the cardio dataset since it failed to converge for the $(\texttt{post-emb}, \texttt{post-enc})$ and $(\texttt{post-emb}, \texttt{post-emb})$ architectures and output NaN values in inference. We used the same hyperparameters for all three architectures. For $(\texttt{post-emb}, \texttt{post-enc})$ and $(\texttt{post-emb}, \texttt{post-emb})$ architectures we report the average over $10$ runs. We display the results in Table \ref{tab:ret_loc}. 
        
        We observe that the ($\texttt{post-enc}$, $\texttt{post-enc}$) architecture obtains the highest mean and lowest mean standard deviation over the tested datasets by a sizable margin. The transformer encoder's expressiveness allows for better representations of a data sample than the embedding layer and may account for such results. Indeed, this shows that the retrieval modules that receive the embedded representation as inputs are less able to select the relevant sample to foster mask reconstruction and anomaly detection performance. Moreover, since the encoder and retrieval modules are trained conjointly, the retrieval module can help the encoder converge to a state that favors sample representations that allow relevant clusters to be formed.
        \begin{table}[t!]
            \begin{center}
                \caption{Comparison of transformer+$\texttt{attention-bsim}$ across values of $\lambda$ based on the F1-Score ($\uparrow$).}
                    \label{tab:rez_lambda}
                    \begin{tabular}{ccccccccc}
                        \toprule
                             $\lambda$ 
                             & $0$ 
                             & $0.1$ 
                             & $0.2$ 
                             & $0.3$ 
                             & $0.4$ 
                             & $0.5$ 
                             & $0.6$ 
                             & $0.7$ \\
                        \midrule
                             Abalone 
                             & $42.5$
                             & $44.3$ 
                             & $47$
                             & $47.3$ 
                             & $46.5$
                             & $\mathbf{53}$  
                             & $46.5$
                             & $45.8$\\ 
                             Satellite
                             & $65.6$
                             & $72$
                             & $72$
                             & $72.5$
                             & $73$
                             & $71.9$
                             & $\mathbf{73.5}$
                             & $73.4$  \\
                             Lympho	
                             & $88.3$
                             & $90.8$
                             & $90.8$
                             & $91.7$
                             & $\mathbf{92.5}$
                             & $90$
                             & $91.7$
                             & $90$ \\
                             Satim.
                             & $89$
                             & $94.5$
                             & $\mathbf{94.6}$
                             & $94.1$
                             & $94.1$
                             & $93.2$ 
                             & $93.8$ 
                             & $94.1$\\
                             Thyroid
                             & $55.5$
                             & $57.2$
                             & $\mathbf{57.4}$
                             & $56.6$
                             & $56.6$
                             & $55.8$
                             & $57.1$
                             & $57.3$ \\
                             Cardio
                             & $68.8$
                             & $\mathbf{80.9}$
                             & $80.6$
                             & $\mathbf{80.9}$
                             & $\mathbf{80.9}$
                             & $\mathbf{80.9}$
                             & $\mathbf{80.9}$
                             & $80.7$ \\
                             Ionosp.
                             & $88.1$
                             & $90.3$
                             & $92.1$
                             & $92.2$
                             & $91.2$
                             & $91.7$
                             & $\mathbf{92.3}$
                             & $91.7$ \\
                             \midrule
                             mean &
                             $71.1$ &
                             $75.7$&
                             $76.4$&
                             $76.5$&
                             $76.4$&
                             $\mathbf{76.6}$
                             & $76.5$
                             & $76.1$ \\
                        \bottomrule
                    \end{tabular}
                \end{center}
            \end{table}
    \subsection{Random mask bank}
        \label{subsec:random_mask}
         
        We also investigate the impact of constructing a random mask bank for inference instead of selecting a deterministic mask bank as discussed in section \ref{subsec:anomalyscore}. We construct for inference a random mask bank composed of the \textit{same number of masks} as for the deterministic mask bank and use the same probability $p_{mask}$ as used to train the model. We compare the performance of the transformer model+$\texttt{attention-bsim}$ based on the F1-score for the two set-ups and display the results in Table \ref{tab:random_masks}. 
        We observe that the deterministic mask bank detects anomalies better on most tested datasets. When computing the anomaly score with the deterministic mask bank, the model obtains an average F1-score of $76.6$ over the $7$ datasets, while with the random mask bank, the model obtains $65.4$. Moreover, as expected, we also observed a significantly larger standard deviation between runs. We might expect the standard deviation to decrease as the number of masks increases, which would induce significant computational overhead.

    \begin{table}[ht!]
            \caption{Comparison of the performance of the transformer + $\texttt{attention-bsim}$ model based on the F1-Score ($\uparrow$) for two mask bank set-ups. The same model was used for inference in both set-ups. We report an average over $10$ different splits of the data.}
            \centering
                \begin{tabular}{ccc}
                \toprule
                     Mask bank & random  & deterministic \\
                \midrule
                    Abalone 
                    & $43.0\pm$\small{$14.9$}
                    & $\mathbf{53.0}\pm$\small{$5.7$}
                    \\
                    Satellite 
                    & $58.4\pm$\small{$5.5$}
                    & $\mathbf{71.9}\pm$\small{$1.5$}
                    \\
                    Lympho
                    & $86.7\pm$\small{$8.5$}
                    & $\mathbf{90.0}\pm$\small{$8.1$}
                    \\
                    Satimage
                    & $47.7\pm$\small{$20.5$}
                    & $\mathbf{93.2}\pm$\small{$1.7$}
                    \\
                    Thyroid
                    & $\mathbf{55.6}\pm$\small{$6.1$}
                    & $\mathbf{55.8}\pm$\small{$6.3$}
                    \\
                    Cardio
                    & $\mathbf{81.3}\pm$\small{$1.6$}
                    & $80.6\pm$\small{$2.4$}
                    \\
                    Ionosphere
                    & $85.2\pm$\small{$4.3$}
                    & $\mathbf{91.7}\pm$\small{$2.1$}
                    \\
                \midrule
                    mean 
                    & $65.4$
                    & $\mathbf{76.6}$ \\
                    mean std
                    & $8.8$
                    & $\mathbf{4.0}$
                \\
                \bottomrule
                \end{tabular}
                \label{tab:random_masks}
        \end{table}
        
    \section{Limitations and Conclusion}
    \label{sec:conclusion}

        \paragraph{Limitations} As with most non-parametric models that leverage the training set in inference, 
        our retrieval-augmented models display a higher complexity than parametric approaches. These approaches can scale well for datasets with a reasonable number of features $d$; however, for large values of $d$, these models incur a high memory cost.

        \paragraph{Conclusion}
        In this work, we have proposed an extensive investigation into external retrieval to augment reconstruction-based anomaly detection methods for tabular data. We have shown that augmenting existing AD methods using attention-based retrieval modules can help foster performance by allowing the model to attend to sample-sample dependencies. 
        Indeed, our experiments involving an extensive benchmark of tabular datasets demonstrate the effectiveness of retrieval-based approaches since the architecture involving the $\texttt{attention-bsim}$ module surpasses the vanilla transformer by a significant margin. We also provide a first explanation as to why combining both types of dependencies can be critical to obtaining consistent performance across datasets: different types of anomalies require different types of dependencies to be efficiently detected.
       
       \paragraph{Future Work} Overall, our work showed that the best-performing attention-based retrieval mechanism relies on other forms of attention than vanilla attention. Parallel to that, models like NPT-AD have shown strong performance for anomaly detection on tabular data and rely on standard multi-head self-attention through the Attention Between Datapoints (ABD) mechanism, close to the $\texttt{v-attention}$ module, to leverage inter-sample relations. Our findings may invite further research on modifying the ABD mechanism involved in NPTs to improve their AD performance. Finally, the use of external retrieval modules proved effective for the task of anomaly detection using mask reconstruction. The proposed external retrieval module could also be easily added to existing deep-AD methods to test whether they may prove relevant for other pretext tasks for anomaly detection on tabular data.

\section*{Acknowledgment}
    This work was granted access to the HPC resources of IDRIS under the allocation 2023-101424 made by GENCI.
    This research publication is supported by the Chair "Artificial intelligence applied to credit card fraud detection and automated trading" led by CentraleSupelec and sponsored by the LUSIS company.

\bibliographystyle{ACM-Reference-Format}
\balance
\bibliography{bibliography}

\onecolumn
\newpage
\appendix 

    \section{Datasets characteristics and experimental settings}
    \label{appendix:datasets}

        \subsection{Dataset characteristics}
        \label{appendix:dataset_char}
            In Table \ref{tab:characteristics}, we display the main characteristics of the datasets involved in our experiments.
            
            \begin{table}[h!]
                \centering
                \caption{Datasets characteristics}
                    \begin{tabular}{cccc}
                        \toprule
                        Dataset &  $n$ & $d$ & Outliers \\
                        \midrule
                        
                        Wine 
                        &$129$ 
                        &$13$
                        & $10$ ($7.7$\%) \\
                        
                        Lympho
                        & $148$
                        & $18$
                        & $6$ ($4.1$\%)\\
                        
                        Glass 
                        &$214$ 
                        &$9$
                        & $9$ ($4.2$\%)\\
                        
                        Vertebral
                        & $240$
                        & $6$
                        & $30$ ($12.5$\%)\\
                        
                        WBC 
                        &$278$
                        & $30$
                        & $21$ ($5.6$\%)\\
                        
                        Ecoli
                        & $336$ 
                        &$7$
                        & $9$ ($2.6$\%)\\
                        
                        Ionosphere
                        & $351$
                        & $33$ 
                        &$126$ ($36$\%)\\
                        
                        Arrhythmia
                        & $452$
                        & $274$ 
                        &$66$ ($15$\%)\\
                        
                        BreastW
                        & $683$
                        & $9$ 
                        &$239$ ($35$\%)\\
                        
                        Pima
                        & $768$
                        & $8$
                        & $268$ ($35$\%)\\
                        
                        Vowels
                        & $1456$
                        & $12$
                        & $50$ ($3.4$\%)\\
                        
                        Letter Recognition
                        & $1600$
                        & $32$
                        & $100$ ($6.25$\%)\\
                        
                        Cardio
                        & $1831$
                        & $21$ 
                        &$176$ ($9.6$\%)\\
                        
                        Seismic
                        & $2584$
                        & $11$
                        & $170$ ($6.5$\%)\\
                        
                        Musk
                        & $3062$
                        & $166$
                        & $97$ ($3.2$\%)\\
                        
                        Speech
                        & $3686$
                        & $400$
                        & $61$ ($1.65$\%)\\
                        
                        Thyroid
                        & $3772$
                        & $6$
                        & $93$ ($2.5$\%)\\
                        
                        Abalone
                        & $4177$
                        & $9$
                        & $29$ ($0.69$\%)\\
                        
                        Optdigits
                        & $5216$
                        & $64$
                        & $150$ ($3$\%)\\
                        
                        Satimage-2
                        & $5803$
                        & $36$
                        & $71$ ($1.2$\%)\\
                        
                        Satellite
                        & $6435$
                        & $36$
                        & $2036$ ($32$\%)\\
                        
                        Pendigits 
                        &$6870$ 
                        &$16$
                        & $156$ ($2.27$\%)\\
                        
                        Annthyroid
                        & $7200$
                        & $6$
                        & $534$ ($7.42$\%)\\
                        
                        Mnist 
                        &$7603$
                        & $100$
                        & $700$ ($9.2$\%)\\
                        
                        Mammography
                        & $11183$
                        & $6$
                        & $260$ ($2.32$\%)\\
                        
                        Shuttle
                        & $49097$
                        & $9$ 
                        &$3511$ ($7$\%)\\
                                                
                        Mulcross
                        & $262144$
                        & $4$
                        & $26214$ ($10$\%)\\
                        
                        ForestCover
                        & $286048$
                        & $10$
                        & $2747$ ($0.9$\%)\\
                                                
                        Campaign
                        & $41188$
                        & $62$ 
                        & $4640$ ($11.3$\%) \\
                        
                        Fraud 
                        & $284807$ 
                        & $29$ 
                        & $492$ ($0.17$\%) 
                        \\

                        Backdoor 
                        & $95329$ 
                        & $196$ 
                        & $2329$ ($2.44$\%) 
                        \\
                        \bottomrule
                    \end{tabular}
                \label{tab:characteristics}
            \end{table}
            \clearpage

    \section{Detailed experiments}
    \label{appendix:results}

    \subsection{Detailed tables for main experiments}
    \label{appendix:detailed_rez}

    \begin{table*}[ht!]
    \begin{center}
        \caption{Anomaly detection F1-score ($\uparrow$). We perform $5$\% T-test to test whether the difference between the highest metrics for each dataset is statistically significant.}
        \label{tab:detailed_table_f1}
    \begin{tabular}{lccccc}
       \toprule
       Method 
        & Transformer 
        & $+\texttt{KNN}$ 
        & $+\texttt{v-att.}$ 
        & $+\texttt{att-bsim}$
        & $+\texttt{att-bsim}$
        \\
     & & & & & $\texttt{-bval}$  \\
       \midrule
       
        Wine 
        & $23.5\pm$\small{$7.9$}
        & $24.9\pm$\small{$5.9$}
        & $26.5\pm$\small{$7.3$}
        & $\mathbf{29.0}\pm$\small{$7.7$}
        & $27.0\pm$\small{$5.6$}
        \\
        
        Lympho 
        & $88.3\pm$\small{$7.6$}
        & $89.2\pm$\small{$7.9$}
        & $88.3\pm$\small{$9.3$}
        & $\mathbf{90.0}\pm$\small{$8.1$}
        & $89.1\pm$\small{$7.9$}
        \\
        
        Glass 
        & $\mathbf{14.4}\pm$\small{$6.1$}
        & $12.8\pm$\small{$3.9$}
        & $12.2\pm$\small{$3.3$}
        & $12.8\pm$\small{$3.9$}
        & $11.1\pm$\small{$0.1$}
        \\
        
        Verteb. 
        & $12.3\pm$\small{$5.2$}
        & $\mathbf{15.7}\pm$\small{$2.8$}
        & $12.7\pm$\small{$4.1$}
        & $14.3\pm$\small{$2.6$}
        & $13.5\pm$\small{$5.1$}
        \\
        
        Wbc 
        & $66.4\pm$\small{$3.2$}
        & $65.2\pm$\small{$4.0$}
        & $66.2\pm$\small{$5.2$}
        & $65.5\pm$\small{$4.4$}
        & $\mathbf{67.9}\pm$\small{$4.2$}
        \\
        
        Ecoli 
        & $75.0\pm$\small{$9.9$}
        & $\mathbf{75.6}\pm$\small{$7.5$}
        & $\mathbf{75.6}\pm$\small{$9.7$}
        & $73.8\pm$\small{$8.0$} 
        & $75.0\pm$\small{$9.7$}
        \\
        
        Ionosph. 
        & $88.1\pm$\small{$2.8$}
        & $85.7\pm$\small{$3.4$}
        & $86.0\pm$\small{$4.7$}
        & $\mathbf{91.7}\pm$\small{$2.1$}
        & $79.3\pm$\small{$1.6$}
        \\
        
        Arrhyth. 
        & $59.8\pm$\small{$2.2$}
        & $60.3\pm$\small{$2.2$}
        & $60.2\pm$\small{$2.7$}
        & $\mathbf{61.2}\pm$\small{$2.1$}
        & $\mathbf{61.1}\pm$\small{$2.8$}
        \\
        
        Breastw 
        & $96.7\pm$\small{$0.3$}
        & $96.7\pm$\small{$0.3$}
        & $\mathbf{96.8}\pm$\small{$0.3$}
        & $96.7\pm$\small{$0.3$}
        & $96.7\pm$\small{$0.3$}
        \\
        
        Pima 
        & $65.6\pm$\small{$2.0$}
        & $64.7\pm$\small{$3.1$}
        & $64.0\pm$\small{$3.3$}
        & $64.3\pm$\small{$2.4$}
        & $\mathbf{67.0}\pm$\small{$1.5$}
        \\
        
        Vowels 
        & $28.7\pm$\small{$8.0$}
        & $40.0\pm$\small{$10.0$}
        & $49.1\pm$\small{$11.1$}
        & $44.5\pm$\small{$10.5$}
        & $\mathbf{58.0}\pm$\small{$11.2$}
        \\
        
        Letter 
        & $41.5\pm$\small{$6.2$}
        & $32.9\pm$\small{$11.8$}
        & $41.8\pm$\small{$11.5$}
        & $\mathbf{43.7}\pm$\small{$10.3$}
        & $28.5\pm$\small{$7.1$}
        \\
        
        Cardio 
        & $\mathbf{68.8}\pm$\small{$2.8$}
        & $65.6\pm$\small{$3.6$}
        & $62.8\pm$\small{$6.9$}
        & $67.7\pm$\small{$3.7$}
        & $68.3\pm$\small{$4.5$}
        \\
        
        Seismic 
        & $\mathbf{19.1}\pm$\small{$5.7$}
        & $17.4\pm$\small{$5.5$}
        & $\mathbf{19.5}\pm$\small{$6.3$}
        & $16.7\pm$\small{$5.5$}
        & $17.5\pm$\small{$5.4$}
        \\
        
        Musk
        & $\mathbf{100}\pm$\small{$0.0$}
        & $\mathbf{100}\pm$\small{$0.0$}
        & $\mathbf{100}\pm$\small{$0.0$}
        & $\mathbf{100}\pm$\small{$0.0$}
        & $\mathbf{100}\pm$\small{$0.0$}
        \\
        
        Speech
        & $\mathbf{6.8}\pm$\small{$1.9$}
        & $6.3\pm$\small{$1.4$}
        & $5.7\pm$\small{$1.7$}
        & $5.9\pm$\small{$1.5$}
        & $5.9\pm$\small{$1.7$}
        \\
        
        Thyroid 
        & $55.5\pm$\small{$4.8$}
        & $55.5\pm$\small{$4.9$}
        & $\mathbf{56.0}\pm$\small{$5.9$}
        & $\mathbf{55.8}\pm$\small{$6.3$}
        & $55.3\pm$\small{$6.6$}
        \\
        
        Abalone 
        & $42.5\pm$\small{$7.8$}
        & $42.5\pm$\small{$9.5$}
        & $49.8\pm$\small{$6.5$}
        & $\mathbf{53.0}\pm$\small{$5.7$}
        & $43.2\pm$\small{$9.1$}
        \\
        
        Optdig. 
        & $61.1\pm$\small{$4.7$}
        & $\mathbf{70.7}\pm$\small{$16.5$}
        & $51.5\pm$\small{$7.6$}
        & $62.6\pm$\small{$6.5$}
        & $22.8\pm$\small{$5$}
        \\
        
        Satimage2
        & $89.0\pm$\small{$4.1$}
        & $86.8\pm$\small{$0.4$}
        & $90.7\pm$\small{$2.6$}
        & $\mathbf{93.2}\pm$\small{$1.7$}
        & $64.2\pm$\small{$7.2$}
        \\
        
        Satellite 
        & $65.6\pm$\small{$3.3$}
        & $58.6\pm$\small{$2.9$}
        & $57.3\pm$\small{$3.0$}
        & $\mathbf{71.9}\pm$\small{$1.5$}
        & $53.7\pm$\small{$3.3$}
        \\
        
        Pendig.
        & $35.4\pm$\small{$10.9$}
        & $\mathbf{52.1}\pm$\small{$9.0$}
        & $39.0\pm$\small{$14.5$}
        & $\mathbf{53.4}\pm$\small{$9.8$}
        & $34.2\pm$\small{$12.2$}
        \\
        
        Annthyr. 
        & $29.9\pm$\small{$1.5$}
        & $\mathbf{30.4}\pm$\small{$1.9$}
        & $\mathbf{30.3}\pm$\small{$1.5$}
        & $\mathbf{30.3}\pm$\small{$1.6$}
        & $\mathbf{30.5}\pm$\small{$1.4$}
        \\
        
        Mnist 
        & $56.7\pm$\small{$5.7$}
        & $\mathbf{64.2}\pm$\small{$3.7$}
        & $61.7\pm$\small{$1.0$}
        & $61.6\pm$\small{$1.0$}
        & $56.7\pm$\small{$1.9$}
        \\
        
        Mammo. 
        & $17.4\pm$\small{$2.2$}
        & $17.3\pm$\small{$2.4$}
        & $15.5\pm$\small{$2.5$}
        & $17.2\pm$\small{$3.0$}
        & $\mathbf{17.7}\pm$\small{$2.7$}
        \\
        
        Shuttle 
        & $85.3\pm$\small{$9.8$}
        & $90.8\pm$\small{$2.9$}
        & $67.7\pm$\small{$13.7$}
        & $87.8\pm$\small{$3.7$}
        & $\mathbf{95.6}\pm$\small{$1.8$}
        \\
        
        Mullcr.
        & $\mathbf{100}\pm$\small{$0.0$}
        & $\mathbf{100}\pm$\small{$0.0$}
        & $\mathbf{100}\pm$\small{$0.0$}
        & $\mathbf{100}\pm$\small{$0.0$}
        & $\mathbf{100}\pm$\small{$0.0$}
        \\
        
        Forest 
        & $21.3\pm$\small{$3.1$}
        & $18.6\pm$\small{$4.6$}	
        & $21.0\pm$\small{$5.9$}
        & $\mathbf{24.9}\pm$\small{$6.5$}
        & $11.1\pm$\small{$4.1$}
        \\
        
        Camp.
        & $47.0\pm$\small{$1.9$}
        & $48.5\pm$\small{$2.1$}
        & $43.3\pm$\small{$2.3$}
        & $49.7\pm$\small{$1.2$}
        & $49.1\pm$\small{$1.1$}
        \\
        
        Fraud 
        & $53.4\pm$\small{$4.4$}
        & $56.4\pm$\small{$2,1$}
        & $56.3\pm$\small{$2.1$}
        & $57.1\pm$\small{$2.1$}
        & $55.2\pm$\small{$1.8$}
        \\
        
        Backd.
        & $85.8\pm$\small{$0.6$}
        & $\mathbf{86.1}\pm$ \small{$0.6$}	
        & $85.2\pm$ \small{$0.7$}
        & $85.3\pm$ \small{$0.6$}
        & $82.4\pm$ \small{$1.3$}
        \\
        
        \midrule 
        
        mean 
        & $56.2$
        & $56.1$
        & $55.7$
        & $\mathbf{58.6}$
        & $53.9$
        \\
        
        \mbox{mean std} 
        & $4.4$
        & $4.6$
        & $5.0$
        & $\mathbf{3.7}$
        & $\mathbf{3.7}$
        \\
        
       \bottomrule
    \end{tabular}
    \end{center}
    \end{table*}

    \begin{table*}[ht!]
        \begin{center}
            \caption{Anomaly detection AUROC($\uparrow$). We perform $5$\% T-test to test whether the differences between the highest metrics for each dataset are statistically significant.}
            \label{tab:detailed_table_auc}
        \begin{tabular}{lccccc}
        \toprule
        Method 
        & Transformer 
        & $+\texttt{KNN}$ 
        & $+\texttt{v-att.}$ 
        & $+\texttt{att-bsim}$
        & $+\texttt{att-bsim}$ \\
        & & & & & $\texttt{-bval}$  \\
       \midrule
           
            Wine 
            & $61.4\pm $\small{$6.7$}
            & $60.4\pm $\small{$5.4$}
            & $62.1\pm $\small{$6.4$}
            & $\mathbf{63.5}\pm $\small{$7.8$}
            & $\mathbf{64.5}\pm $\small{$5.6$}
            \\
            Lympho 
            & $99.5\pm $\small{$0.4$}
            & $99.6\pm $\small{$0.4$}
            & $99.6\pm $\small{$0.5$}
            & $\mathbf{99.7}\pm $\small{$0.3$}
            & $\mathbf{99.7}\pm $\small{$0.3$}
            \\
            Glass 
            & $61.2\pm $\small{$7.0$}
            & $61.2\pm $\small{$5.0$}
            & $\mathbf{62.1}\pm $\small{$7.0$}
            & $59.3\pm $\small{$6.9$}
            & $59.1\pm $\small{$5.8$}
            \\
            Vertebral
            & $44.8\pm $\small{$5.2$}
            & $\mathbf{46.7}\pm $\small{$4.1$}
            & $45.3\pm $\small{$7.1$}
            & $45.4\pm $\small{$3.7$} 
            & $45.4\pm $\small{$4.7$}
            \\
            WBC 
            & $95.0\pm $\small{$1.1$}
            & $94.3\pm $\small{$1.5$}
            & $94.3\pm $\small{$1.6$}
            & $94.2\pm $\small{$1.1$}
            & $\mathbf{95.5}\pm $\small{$1.6$}
            \\
            Ecoli 
            & $84.8\pm $\small{$1.6$}
            & $84.8\pm $\small{$1.8$} 
            & $85.2\pm $\small{$2.7$}
            & $\mathbf{87.4}\pm $\small{$1.8$}
            & $85.4\pm $\small{$2.3$}
            \\
            Ionosph. 
            & $95.4\pm $\small{$1.9$}
            & $93.7\pm $\small{$2.7$}
            & $93.6\pm $\small{$4.0$}
            & $\mathbf{97.5}\pm $\small{$0.1$}
            & $87.2\pm $\small{$2.3$}
            \\
            Arrhyth.
            & $81.7\pm$\small{$1.1$} 
            & $81.9\pm$\small{$0.9$} 
            & $81.8\pm$\small{$0.9$} 
            & $\mathbf{82.3}\pm$\small{$0.7$} 
            & $\mathbf{82.1}\pm$\small{$0.9$} 
            \\
            Breastw 
            & $\mathbf{99.6}\pm$\small{$0.1$}
            & $\mathbf{99.6}\pm$\small{$0.1$}
            & $\mathbf{99.6}\pm$\small{$0.1$}
            & $\mathbf{99.6}\pm$\small{$0.1$}
            & $\mathbf{99.6}\pm$\small{$0.1$}
            \\
            Pima
            & $67.2\pm$\small{$2.4$} 
            & $66.0\pm$\small{$3.8$} 
            & $65.4\pm$\small{$3.8$} 
            & $65.8\pm$\small{$2.9$} 
            & $\mathbf{68.7}\pm$\small{$1.4$} 
            \\
            
            Vowels
            & $78.4\pm$\small{$9.2$} 
            & $86.1\pm$\small{$5.2$} 
            & $90.4\pm$\small{$4.7$} 
            & $88.3\pm$\small{$4.5$} 
            & $\mathbf{94.3}\pm$\small{$2.8$} 
            \\
            
            Letter
            & $80.5\pm$\small{$4.8$} 
            & $73.5\pm$\small{$9.6$} 
            & $\mathbf{81.0}\pm$\small{$8.7$} 
            & $\mathbf{81.5}\pm$\small{$6.8$} 
            & $69.1\pm$\small{$7.7$} 
            \\
            Cardio
            & $\mathbf{93.5}\pm$\small{$1.3$} 
            & $92.0\pm$\small{$1.7$} 
            & $89.9\pm$\small{$4.2$} 
            & $\mathbf{93.3}\pm$\small{$1.7$} 
            & $\mathbf{93.7}\pm$\small{$1.3$} 
            \\
            
            Seismic
            & $\mathbf{58.2}\pm$\small{$7.9$} 
            & $56.8\pm$\small{$8.4$} 
            & $57.9\pm$\small{$7.6$} 
            & $\mathbf{58.0}\pm$\small{$6.7$} 
            & $54.8\pm$\small{$6.2$} 
            \\
            
            Musk
            & $\mathbf{100}\pm$\small{$0.0$}
            & $\mathbf{100}\pm$\small{$0.0$}
            & $\mathbf{100}\pm$\small{$0.0$}
            & $\mathbf{100}\pm$\small{$0.0$}
            & $\mathbf{100}\pm$\small{$0.0$}
            \\
            
            Speech
            & $47.2\pm$\small{$0.7$}
            & $\mathbf{47.3}\pm$\small{$0.8$}
            & $\mathbf{47.3}\pm$\small{$0.8$}
            & $\mathbf{47.3}\pm$\small{$0.8$}
            & $47.0\pm$\small{$0.5$}
            \\
            
            Thyroid
            & $\mathbf{93.8}\pm$\small{$1.2$} 
            & $\mathbf{93.8}\pm$\small{$1.2$} 
            & $93.6\pm$\small{$5.9$} 
            & $93.7\pm$\small{$1.5$} 
            & $93.6\pm$\small{$1.8$} 
            \\
            Abalone
            & $\mathbf{88.3}\pm$\small{$2.0$} 
            & $86.1\pm$\small{$3.6$} 
            & $\mathbf{88.0}\pm$\small{$3.5$} 
            & $\mathbf{87.9}\pm$\small{$3.7$} 
            & $86.6\pm$\small{$2.9$} 
            \\
            Optdig.
            & $96.4\pm$\small{$4.7$}
            & $96.2\pm$\small{$9.8$}
            & $94.9\pm$\small{$1.7$}
            & $\mathbf{96.7}\pm$\small{$1.1$}
            & $83.4\pm$\small{$3.2$}
            \\
            Satimage
            & $99.7\pm$\small{$0.1$}
            & $99.5\pm$\small{$0.2$}
            & $99.6\pm$\small{$0.2$}
            & $\mathbf{99.8}\pm$\small{$0.1$}
            & $96.8\pm$\small{$1.9$}
            \\
            Satellite
            & $73.8\pm$\small{$2.5$}
            & $68.9\pm$\small{$2.0$}
            & $67.8\pm$\small{$2.5$}
            & $\mathbf{79.5}\pm$\small{$1.9$}
            & $62.0\pm$\small{$2.9$}
            \\
            Pendigits
            & $93.8\pm$\small{$2.6$}
            & $96.5\pm$\small{$1.4$}
            & $94.1\pm$\small{$2.8$}
            & $\mathbf{97.1}\pm$\small{$1.2$}
            & $89.4\pm$\small{$7.1$}
            \\
            Annthyr.
            & $65.4\pm$\small{$1.4$}
            & $66.0\pm$\small{$1.7$}
            & $\mathbf{66.2}\pm$\small{$1.3$}
            & $\mathbf{66.2}\pm$\small{$1.6$}
            & $66.0\pm$\small{$1.1$}
            \\
            Mnist 
            & $87.4\pm$\small{$3.2$}
            & $\mathbf{90.3}\pm$\small{$2.2$}
            & $89.9\pm$\small{$0.4$}
            & $\mathbf{90.0}\pm$\small{$0.5$}
            & $87.3\pm$\small{$1.0$}
            \\
            Mammo.
            & $77.6\pm$\small{$1.0$}
            & $76.8\pm$\small{$2.4$}
            & $75.2\pm$\small{$2.9$}
            & $78.4\pm$\small{$1.6$}
            & $\mathbf{79.8}\pm$\small{$1.8$}
            \\
            Mullcross
            & $\mathbf{100}\pm$\small{$0.0$}
            & $\mathbf{100}\pm$\small{$0.0$}
            & $\mathbf{100}\pm$\small{$0.0$}
            & $\mathbf{100}\pm$\small{$0.0$}
            & $\mathbf{100}\pm$\small{$0.0$}
            \\
            Shuttle
            & $97.2\pm$\small{$2.2$}
            & $98.1\pm$\small{$0.5$}
            & $90.2\pm$\small{$6.1$}
            & $97.7\pm$\small{$0.9$}
            & $\mathbf{98.9}\pm$\small{$0.3$}
            \\
            
            Forest
            & $95.1\pm$\small{$0.8$} 
            & $94.5\pm$\small{$0.7$}
            & $95.0\pm$\small{$1.0$}
            & $\mathbf{95.4}\pm$\small{$0.9$}
            & $92.7\pm$\small{$0.7$}
            \\
            
            Campaign
            & $75.3\pm$\small{$2.1$}
            & $75.7\pm$\small{$1.9$}
            & $69.3\pm$\small{$2.0$}
            & $76.1\pm$\small{$2.0$}
            & $75.9\pm$\small{$2.1$}
            \\
            Fraud
            & $94.7\pm$\small{$0.4$}
            & $95.1\pm$\small{$0.4$}
            & $95.2\pm$\small{$0.4$}
            & $\mathbf{95.8}\pm$\small{$0.4$}
            & $94.7\pm$\small{$0.4$}
            \\
            Backdoor
            & $\mathbf{95.1}\pm$\small{$0.2$}
            & $\mathbf{95.2}\pm$\small{$0.3$}
            & $94.5\pm$\small{$0.2$}
            & $94.7\pm$\small{$0.1$}
            & $91.7\pm$\small{$0.3$}
            \\
            
            \midrule
            
            mean 
            & $83.4$
            & $83.1$
            & $83.1$
            & $\mathbf{84.4}$
            & $82.1$
            \\
            
            \mbox{mean std} 
            & $2.4$
            & $2.8$
            & $2.6$
            & $2.0$
            & $2.3$
            \\

        \bottomrule
        \end{tabular}
        \end{center}
        \end{table*}
    \clearpage

    \subsection{Synthetic dataset generation and experimental detail for section \ref{sec:theoretical_motiv}}
        \label{app:exp_synthetic_dataset}

        The synthetic three dimensional dataset was generated as follows. 
        \begin{itemize}
            \item \textbf{Normal samples}: We consider an interval of values $[-2,3]$ from which we uniformly sample the first feature $x_1$. We then sample $x_2, x_3$ following the relation given in eq. \eqref{eq:feature_relation} with parameters, $(\alpha_1, \beta_1)=(2,3)$, $(\alpha_2, \beta_2)=(4,3)$ and $\varepsilon \sim \mathcal{N}(0,1)$.
            \item \textbf{Anomalies (type 1)}: We consider an interval of values $[3.3,4]$ from which we uniformly sample the first feature $x_1$ and keep the rest as for normal samples.
            \item \textbf{Anomalies (type 2)}: We consider an interval of values $[1.5,2.5]$ from which we uniformly sample the first feature $x_1$ and sample $x_2, x_3$ following eq. \eqref{eq:feature_relation} but with parameters $(\alpha_1, \beta_1)=(-7.5,-1)$ and $(\alpha_2, \beta_2)=(4,3)$. 
        \end{itemize}
        The vanilla transformer and its augmented version were trained with the following hyperparameters:
        \begin{itemize}
            \item Batch size: $-1$.
            \item Patience: $100$ epochs.
            \item Learning rate (lr): $0.001$.
            \item Hidden dim ($e$): $8$.
            \item Masking probability $p_{mask}$: $0.15$.
            \item Number of attention heads: $2$.
            \item Number of layers of the encoder: $2$.
            \item Retrieval hyper-parameters:
                \begin{itemize}
                    \item Retrieval location: post-encoder
                    \item Retrieval aggregation location: post-encoder
                    \item $\lambda$: 0.5
                    \item $\mathcal{C}=\mathcal{D}_{train}$
                    \item $card(\mathcal{H})=500$
                \end{itemize}

        \end{itemize}
        For Mask-KNN, following \cite{thimonier2024beyond} we set the number of neighbors to $k=5$. 
\end{document}